\documentclass[11pt,letterpaper]{article}

\usepackage{cogsys}
\usepackage{natbib}
\usepackage[T1]{fontenc}
\usepackage{times}
\usepackage[pdftex]{graphicx}

\usepackage{url}
\usepackage{hyperref}
\usepackage{xcolor}
\usepackage{listings}
\usepackage{caption}
\usepackage{booktabs}
\usepackage{multirow}
\usepackage{algorithm}
\usepackage{algpseudocode}
\usepackage{wrapfig}
\usepackage{setspace}
\usepackage{float}  
\usepackage[section]{placeins}
\usepackage{soul}        
\usepackage{upquote}       
\usepackage{inconsolata}   
\usepackage{amssymb}

\DeclareCaptionFont{white}{\color{white}}
\DeclareCaptionFormat{listing}{%
  \parbox{\linewidth}{\colorbox{gray}{\parbox{\linewidth}{#1#2#3}}\vskip-4pt}}
\usepackage{amsmath,amsfonts,bm}

\def\eqref#1{equation~\ref{#1}}

\def\1{\bm{1}}

\DeclareMathAlphabet{\mathsfit}{\encodingdefault}{\sfdefault}{m}{sl}
\SetMathAlphabet{\mathsfit}{bold}{\encodingdefault}{\sfdefault}{bx}{n}

\def\gM{{\mathcal{M}}}

\def\gP{{\mathcal{P}}}

\ShortHeadings{Deliberate Planning in Language Models with Symbolic Representation}{Xiong, Liu, Zhou, and Su}

\cogsysheading{12}{2025}{1--22}{8/2025}{10/2025}

\begin{document}

\title{Deliberate Planning in Language Models with Symbolic Representation}

\author{Siheng Xiong}{sxiong45@gatech.edu}
\author{Zhangding Liu}{zliu952@gatech.edu}
\author{Jieyu Zhou}{jzhou625@gatech.edu}
\author{Yusen Su}{ysu349@gatech.edu}
\address{Georgia Institute of Technology, Atlanta, GA 30332 USA}

\vskip 0.2in

\begin{abstract}
Planning remains a core challenge for large language models (LLMs), particularly in domains that require coherent multi-step action sequences grounded in external constraints.  We introduce \textbf{SymPlanner}, a novel framework that equips LLMs with structured planning capabilities by interfacing them with a symbolic environment that serves as an explicit world model.  Rather than relying purely on natural language reasoning, SymPlanner grounds the planning process in a symbolic state space, where a policy model proposes actions and a symbolic environment deterministically executes and verifies their effects.  To enhance exploration and improve robustness, we introduce \textbf{Iterative Correction (IC)}, which refines previously proposed actions by leveraging feedback from the symbolic environment to eliminate invalid decisions and guide the model toward valid alternatives.  Additionally, \textbf{Contrastive Ranking (CR)} enables fine-grained comparison of candidate plans by evaluating them jointly.  
Conceptually, SymPlanner operationalizes two cognitive faculties: (i) error monitoring and repair via externalized feedback (IC) and (ii) preference formation among alternatives via pairwise comparison (CR), advancing cognitively plausible, symbol-grounded planning aligned with the rich structure in intelligent systems.
We evaluate SymPlanner on PlanBench, demonstrating that it produces more coherent, diverse, and verifiable plans than pure natural language baselines\footnote{Code and data are available at \url{https://github.com/xiongsiheng/SymPlanner}.}.
\end{abstract}

\section{Introduction}

Planning is a fundamental aspect of intelligent behavior, requiring the ability to construct coherent action sequences that achieve specific goals while respecting environmental constraints. In real-world applications such as robotics, virtual agents, and interactive assistants \citep{jiang2024towards,jin2024agentreview,liu2025examining}, effective planning demands not only long-horizon decision-making but also strict adherence to logical consistency and domain-specific rules. For instance, a household robot tasked with rearranging objects must ensure that every intermediate step, such as lifting, placing, or stacking, remains valid according to physical constraints. Similarly, virtual assistants must generate interaction strategies that align with user preferences and task requirements without violating contextual limitations. These scenarios illustrate the dual challenge of maintaining both local validity (i.e., ensuring individual actions are executable) and global coherence (i.e., ensuring the sequence as a whole leads toward the intended goal).
While language models (LMs) have demonstrated impressive reasoning and problem-solving capabilities, they continue to struggle in planning tasks where actions must be grounded in a verifiable and coherent world model \citep{valmeekam2023planbench}. Their reliance on natural language reasoning alone often leads to error-prone outputs: action sequences that appear plausible in text but fail under precise execution. This challenge becomes increasingly severe in domains that require structured, multi-step decision-making under rigid constraints, as early mistakes propagate and accumulate, resulting in incoherent or infeasible plans.

\begin{figure}[t]
\centering
\includegraphics[width=\linewidth]{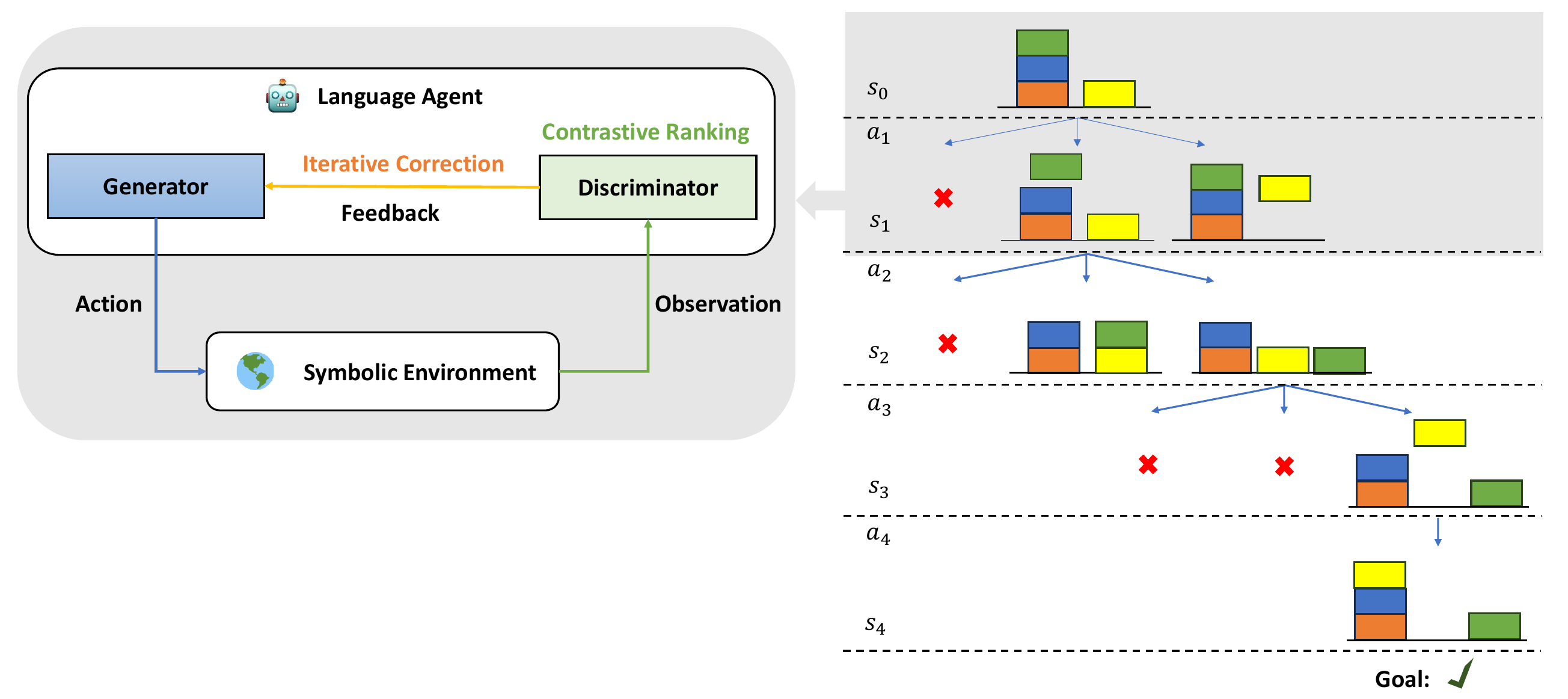}
\caption{Overview of SymPlanner. The language agent consists of a generator and a discriminator, which interact with a symbolic environment. The generator proposes candidate actions, which are refined via \textit{iterative correction} using feedback from the environment. The discriminator then applies \textit{contrastive ranking} over simulated outcomes to select the most coherent successor state. On the right, an example blocksworld task illustrates how invalid actions are pruned (red crosses) and valid trajectories are retained until the goal configuration is achieved (green check).}
\label{fig:framework}
\end{figure}

Existing LLM-based planning paradigms, such as chain-of-thought (CoT) reasoning \citep{wei2022chain}, tree-of-thought (ToT) exploration \citep{yao2023tree}, and reasoning-as-planning (RAP) \citep{hao2023reasoning}, primarily rely on natural language to represent both states and actions. This design provides flexibility and ease of integration, as natural language serves as a universal interface for reasoning across domains. However, it also introduces several critical limitations. First, natural language descriptions are inherently ambiguous and can be interpreted in multiple ways \citep{jiang2024peek}, making it difficult for models to guarantee precise alignment between a described action and its actual effect. Second, these paradigms often lack mechanisms for validation: without grounding in an explicit environment, models cannot verify whether an action is executable, nor can they ensure that a sequence of actions consistently advances toward the goal. Third, as task complexity and planning horizons increase, small inconsistencies in language-based reasoning tend to accumulate, leading to incoherent or infeasible plans \citep{yang2024can}. For example, a model may generate a sequence that instructs stacking a block onto another that is not yet cleared, or it may repeat contradictory actions such as picking up the same object twice without putting it down. These errors are symptomatic of the gap between high-level linguistic reasoning and low-level, rule-governed execution. Consequently, while language-based approaches excel at producing diverse strategies and human-readable rationales, they often fall short in domains that demand precise, verifiable, and logically consistent multi-step planning.

To address these limitations, we propose SymPlanner, a novel framework that augments language models with structured planning capabilities through integration with a symbolic environment. The symbolic environment serves as an explicit and deterministic world model, enabling actions to be executed and validated against clearly defined rules. This grounding mechanism ensures that every decision is checked for feasibility and coherence, thereby reducing the risk of cascading errors. Within SymPlanner, a policy model generates candidate actions, which are executed and verified by the symbolic world model. When an action is found invalid, the system leverages Iterative Correction (IC) to refine it using symbolic feedback, guiding the LLM toward valid alternatives rather than discarding its output. Furthermore, we introduce Contrastive Ranking (CR), which enables the discriminator to evaluate multiple candidate plans jointly rather than independently, providing fine-grained comparisons that improve plan selection quality. Together, these mechanisms enhance exploration, increase robustness, and support more deliberate and reliable planning.
We evaluate SymPlanner on PlanBench, a challenging benchmark for plan generation and verification, where agents manipulate blocks in structured environments. Experimental results demonstrate that SymPlanner produces plans that are significantly more valid, coherent, and diverse than natural language-only baselines. Unlike prior methods, which deteriorate quickly as task horizons grow, SymPlanner maintains competitive success rate even for longer sequences, highlighting its scalability and robustness in multi-step reasoning tasks.

Our contributions can be summarized as follows:
\begin{itemize}
\item We introduce SymPlanner, a framework that couples language models with a symbolic environment, providing structured and verifiable planning capabilities that address the limitations of purely natural language-based reasoning.
\item We propose two mechanisms: Iterative Correction, a feedback-driven refinement process for invalid actions, and Contrastive Ranking, a joint evaluation strategy for candidate plans, that together improve robustness and decision quality.
\item We conduct comprehensive experiments on PlanBench, demonstrating that SymPlanner outperforms natural language-only baselines in terms of planning success rate, coherence, and execution fidelity, particularly in long-horizon scenarios.
\end{itemize}

\section{Methodology}
\label{sec:method}

We introduce \textbf{SymPlanner}, a framework that augments language models (LMs) with structured planning capabilities through integration with a \textbf{symbolic environment}. The framework combines three components: (i) a policy model $\gP_{\pi}$ that generates candidate actions, (ii) a discriminator $\gP_{\text{d}}$ that evaluates and ranks candidate trajectories, and (iii) a symbolic world model $\gM_{\text{wm}}$ that deterministically simulates the effects of actions and enforces domain constraints. Together, these components allow LLMs to reason in a symbolic state space, reducing ambiguity and ensuring that generated plans are both coherent and verifiable.

\subsection{Problem definition}
\label{sec:problem}

We model a symbolic planning problem as a tuple
\(\mathcal{P}=(\mathcal{F},\mathcal{A}, I, G)\) \citep{ghallab2004automated, geffner2013concise}.
\(\mathcal{F}\) is a set of ground atoms over typed predicates;
\(\mathcal{A}\) is a set of grounded actions with preconditions \(\mathrm{Pre}(a)\subseteq \mathcal{F}\) and add/delete effects \(\mathrm{Add}(a),\mathrm{Del}(a)\subseteq \mathcal{F}\).
A (symbolic) state is a set \(s\subseteq \mathcal{F}\). An action \(a\) is applicable in \(s\) iff \(\mathrm{Pre}(a)\subseteq s\), and the transition function is
\(\gamma(s,a) = (s \setminus \mathrm{Del}(a)) \cup \mathrm{Add}(a)\).
The initial state is \(I\); the goal \(G\subseteq \mathcal{F}\) is a partial state and is satisfied by \(s\) if \(s \models G\) (i.e., \(G\subseteq s\)).

Given a sequence (plan) \(\Pi=[a_1,\dots,a_T]\), we write \(\gamma(I,\Pi)\) for the result of successively applying \(\gamma\). Our objective is to find a plan with
\[
\gamma(I,\Pi) \models G, \quad \text{minimizing cost (here, plan length).}
\]

In SymPlanner, an LLM proposes candidate actions conditioned on the current \emph{symbolic} state; a symbolic simulator serves as the world model and the arbiter of validity. Natural‑language goals \(g^{\text{NL}}\) and states (if provided) are mapped to symbols by a parser \(\phi_{\text{NL}\to\text{SYM}}\), yielding \(G=\phi(g^{\text{NL}})\) and \(I=\phi(s^{\text{NL}})\). All feasibility checks and transitions operate over symbols.

\subsection{System design}

The core idea of SymPlanner is to structure the planning process as an
\emph{iterative search in symbolic state space}, guided by the generative flexibility
of LLMs and constrained by the deterministic rules of a world model.
Rather than relying on a single roll-out, the system maintains two explicit pools:
a \emph{completed pool} $\mathcal{D}$, which stores plans that already reach the goal,
and an \emph{unfinished pool} $\mathcal{C}$, which contains partial trajectories that
can be further expanded. This design allows the system to balance \emph{exploration}
(through unfinished trajectories) with \emph{exploitation} (by recording and ranking
completed plans).

\paragraph{Action proposal (generation).}
Given current symbolic state $s$ and goal $G$, the policy LLM $\gP_{\pi}$
proposes $N$ candidate actions
\[
\mathcal{A}=\{\,a^{(i)} \sim \gP_{\pi}(\cdot \mid s,G,\Pi)\,\}_{i=1}^{N},
\]
where $\Pi$ is the partial plan used as context. Parameters are not updated at test time.

\paragraph{Symbolic simulation (validation \& transition).}
The world model $\gM_{\text{wm}}$ (VAL-compatible simulator) is the arbiter of validity:
\[
\texttt{step}(s,a) \;=\;
\begin{cases}
\gamma(s,a), & \text{if } \mathrm{Pre}(a)\subseteq s,\\
\bot,\,\varepsilon, & \text{otherwise,}
\end{cases}
\]
where $\varepsilon$ is a typed error (e.g., \texttt{HandNotEmpty}, \texttt{NotClear}$(x)$).
We prefer an explicit failure signal $\bot$ over a no-op so that invalid proposals trigger repair (IC) rather than silently stalling search.

\paragraph{Budgets and widths.}
We distinguish two key parameters: (i) \emph{beam width} $b$, the number of successors retained at each layer, and (ii) \emph{rollout budget} $N$, the number of candidate actions sampled for each successor.
We maintain two pools throughout the process: \textsc{Open}, representing the frontier of partial plans to be expanded, and \textsc{Done}, storing successfully completed plans.

\begin{algorithm}[t]
\caption{SymPlanner $ (g, s_0, \gP_{\pi}, \gM_{\text{wm}}, \gP_{\text{d}}, N, T, b) $}
\label{alg}
\begin{algorithmic}[1]
\Require Goal $g$, initial state $s_0$, policy $\gP_{\pi}$, world model $\gM_{\text{wm}}$, discriminator $\gP_{\text{d}}$, rollout budget $N$, step limit $T$, beam width $b$
\State $\mathcal{D} \gets \{\}$ \Comment{initialize completed plan pool}
\State $\mathcal{C} \gets \{ (g, s_0, [\,]) \}$ \Comment{initialize unfinished pool with empty plan}
\For{$t=1,\dots,T$} \Comment{iterate for at most $T$ steps}
  \If{$b=0$} \State \textbf{break} \Comment{stop if rollout budget is exhausted} \EndIf
  \State $\mathcal{C} \gets \{ (g, s_{t-1}, \Pi_{t-1}\cup[a_t]) \mid a_t \in \texttt{gen}(\gP_{\pi},(g,s_{t-1}),N)\}$ \Comment{expand each unfinished plan}
  \State $\mathcal{C} \gets \texttt{IC}(\mathcal{C}, \gM_{\text{wm}}, \gP_{\pi})$ \Comment{repair invalid actions with iterative correction}
  \State $\mathcal{C} \gets \texttt{CR}(\gP_{\text{d}}, \{\texttt{symSim}(g,s_{t-1},a_t,T)\}, b)$ \Comment{contrastive ranking over candidates}
  \For{$(g,s_{t-1},\Pi_t) \in \mathcal{C}$} \Comment{evaluate each surviving candidate}
    \If{$s_t=g$} 
      \State $\mathcal{D}.\mathrm{add}(\Pi_t)$ \Comment{goal reached: move plan to completed pool}
      \State $\mathcal{C}.\mathrm{pop}((g,s_{t-1},\Pi_t))$
      \State $b \gets b - 1$ \Comment{reduce beam width}
    \Else 
      \State $\mathcal{C}.\mathrm{add}((g,s_t,\Pi_t))$ \Comment{otherwise keep plan for further expansion}
    \EndIf
  \EndFor
\EndFor
\State \Return $\texttt{CR}(\gP_{\text{d}},\mathcal{D},1)$ \Comment{return best plan from completed pool}
\end{algorithmic}
\end{algorithm}

\paragraph{Planning process.}
Algorithm~\ref{alg} summarizes the workflow. Initially, $\mathcal{C}$ contains the
empty plan with the starting state $s_0$. At each step, the policy $\gP_{\pi}$
generates $N$ candidate actions conditioned on the current goal and state. These
candidates are executed in the symbolic world model $\gM_{\text{wm}}$, which serves
two purposes: (i) it enforces domain constraints by rejecting invalid actions, and
(ii) it deterministically predicts successor states for valid ones. The resulting
state--action pairs are then scored by the discriminator $\gP_{\text{d}}$, which
selects the top $b$ for further expansion. This acts as a \emph{beam search} over
symbolic states, ensuring that the system focuses computational resources on the
most promising branches. 
If a candidate trajectory successfully reaches the goal $g$, its plan $\Pi_t$ is
added to the completed pool $\mathcal{D}$. Otherwise, it is returned to the
unfinished pool for further expansion in subsequent iterations. The process
terminates when the step limit $T$ is reached or the rollout budget $b$ is
exhausted. Finally, among all completed plans, the discriminator selects the best
solution $\Pi^*$, which represents the system's final output. This structure
ensures that SymPlanner can produce plans that are not only coherent in natural
language but also \emph{grounded, verifiable, and consistent} with symbolic rules.

\paragraph{Why not just a classical planner?}
Classical planners excel when a complete domain model and a symbolic goal are given. Our setting targets three practical gaps: (i) goals or intermediate constraints arrive in natural language and must be grounded; (ii) domain models may be incomplete, LLMs propose plausible actions while the simulator enforces safety; (iii) generalization across task variants with new objects/relations where hand‑tuning models is costly. SymPlanner exploits LLMs for proposal and uses the symbolic world model for verification, yielding a safe, language‑grounded planner.

\subsection{Iterative Correction (IC)}
\label{sec:ic}

A common failure mode in LLM-based planning is proposing actions whose
preconditions are not satisfied (e.g., attempting to stack onto a non-clear block).
Rather than discarding such proposals, IC uses the simulator’s typed error to
repair them.

\paragraph{Mechanism.}
Given $(s,a)$ with $\text{Sim.step}(s,a)=\bot$ and error category
$\varepsilon$ (e.g., \texttt{HandNotEmpty}, \texttt{NotClear}$(x)$),
IC regenerates a corrected action conditioned on $\varepsilon$:
\[
a' \sim \gP_{\pi}\big(\cdot \mid s, G, \Pi, \varepsilon\big)
\]
repeating up to $R$ times until a valid action is found
(i.e., $\text{Sim.step}(s,a') \neq \bot$). If no valid action is obtained,
the branch is pruned. Using an explicit failure signal ($\bot,\varepsilon$) instead
of a silent no-op ensures repair is triggered rather than allowing the search to
stall. This feedback reduces error propagation and improves long-horizon stability
\citep{tong2024can, xu2024sayself}.

\subsection{Contrastive Ranking (CR)}
\label{sec:cr}

IC ensures candidates are valid; CR then chooses which valid successors to
expand. Instead of scoring each option independently, CR compares candidates
\emph{relatively} via pairwise preferences, yielding more robust selection.

\paragraph{Mechanism.}
Let $\mathcal{C}=\{(s_i,\Pi_i)\}_{i=1}^m$ be valid successors after simulation.
The discriminator $\gP_{\mathrm d}$ provides pairwise preference logits
(or probabilities) for $(i \succ j)$, conditioned on the goal:
\[
\ell_{ij} \;=\; \gP_{\mathrm d}\!\big((s_i,\Pi_i)\succ(s_j,\Pi_j)\,\big|\,G\big), 
\quad \ell_{ji}=1-\ell_{ij}.
\]
Aggregate a tournament-style score
\[
\mathrm{score}(i)\;=\; \sum_{j\neq i} \mathbb{E}\!\left[\mathbf{1}\{i \succ j\}\right] 
\;\approx\; \sum_{j\in \mathcal{N}_k(i)} \ell_{ij},
\]
where $\mathcal{N}_k(i)$ is a set of $k$ opponents sampled for efficiency,
giving total complexity $O(mk)$. CR returns the $\operatorname{TopK}$
candidates by $\mathrm{score}(\cdot)$:
\[
\texttt{CR}(\mathcal{C},b)\;=\;\operatorname{TopK}\!\Bigl(\mathcal{C},\,\{\mathrm{score}(i)\}_{i=1}^m,\,b\Bigr).
\]
Compared to independent scoring, pairwise contrast stabilizes preferences and
reduces selection of superficially plausible but globally inferior branches
\citep{xiong2024deliberate}.

\section{Experiments}

\subsection{Experimental Setup}

\paragraph{Benchmark.}
We evaluate SymPlanner on PlanBench \citep{valmeekam2023planbench}, a symbolic planning benchmark where a robotic agent manipulates blocks on a table to achieve a target configuration. PlanBench is particularly suitable for our study because it requires strict adherence to symbolic constraints while supporting long-horizon reasoning. The action space consists of four operations (\textit{Pickup}, \textit{Putdown}, \textit{Stack}, and \textit{Unstack}), which are represented and parsed by language models \citep{yang2023harnessing}.

\paragraph{Baselines.}
We compare SymPlanner with three widely used LLM-based reasoning paradigms: chain-of-thought (CoT), tree-of-thought (ToT), and reasoning-as-planning (RAP), implemented via the LLM Reasoners library \citep{hao2024llm}. For CoT, we generate independent reasoning chains with a fixed temperature of $0$. ToT is configured with a generation limit of $4$, a breadth of $4$, and a search depth of $6$, resulting in $32$ total rollouts. RAP is set with a depth limit of $6$, cumulative reward aggregation, mean-based Q-value estimation, and uses Monte Carlo tree search (MCTS) with symbolic feedback. Verification for all methods is performed with a symbolic validator \citep{howey2004val}.

\paragraph{Implementation details.}
For SymPlanner, the maximum number of planning steps is set to $16$, with $32$ rollouts per instance. Each planning episode involves three iterations of generation, and both the group size and beam width are fixed at $3$. For contrastive ranking, each candidate is compared against one alternative (i.e., one pairwise comparison per option). For ToT, RAP, and SymPlanner, the sampling temperature is set to $0.7$. Experiments are conducted with three base models: GPT-4o-mini, GPT-4o, and GPT-4.1. To balance evaluation quality and computational cost, we randomly select $20$ test instances for each step count.

{\small
\begin{table}[t]
\centering
\caption{Success rate (\%) on PlanBench by optimal plan length. A plan counts as a success only if the simulator validates all steps and the final state entails the goal. Mean over 20 instances per length; 95\% CIs in parentheses. Plan validity is checked using symbolic evaluation. Bold numbers indicate the best result for each model.}
\label{tab:main-results}
\resizebox{0.85\textwidth}{!}{%
\small
\begin{tabular}{ll|ccccccc}
\hline
\specialrule{0em}{1pt}{1pt}
\multirow{2}{*}{\textbf{Model}} & \multirow{2}{*}{\textbf{Method}} & \multicolumn{7}{c}{\textbf{Step Count}}\\
\cmidrule(lr){3-9}
& & 2 & 4 & 6 & 8 & 10 & 12 & Total\\
\hline
\specialrule{0em}{1pt}{1pt}
\multirow{5}{*}{GPT-4o-mini} 
& Zero-shot CoT & 5.0 & 0.0 & 5.0 & 0.0 & 0.0 & 0.0 & 1.7 \\
& Few-shot CoT {(4-shot)} & 10.0 & 10.0 & 10.0 & 5.0 & \textbf{5.0} & 0.0 & 6.7 \\
& ToT & 20.0 &20.0  &0.0 &0.0 &0.0 &0.0 &6.7\\
& RAP & \textbf{65.0} &10.0 &0.0 &0.0 &0.0 &0.0 &12.5\\
& SymPlanner & 60.0 & \textbf{45.0} & \textbf{10.0}  & \textbf{10.0} & 0.0 & 0.0 & \textbf{21.6} \\
\hline
\specialrule{0em}{1pt}{1pt}
\multirow{5}{*}{GPT-4o}
& Zero-shot CoT & 40.0 & 20.0 & 15.0 & 10.0 & 20.0 & 0.0 & 17.5 \\
& Few-shot CoT {(4-shot)} & 40.0 & 30.0 & 15.0 & 10.0 & 10.0 & 0.0 & 17.5 \\
& ToT & 35.0  & 15.0  & 5.0  & 0.0  & 0.0  & 0.0 & 9.2 \\
& RAP & 65.0 &40.0 &0.0 &0.0 &0.0 &0.0 &17.5\\
& SymPlanner & \textbf{85.0} & \textbf{45.0} & \textbf{50.0} & \textbf{55.0} & \textbf{45.0} & \textbf{20.0} & \textbf{50.0}\\
\hline
\specialrule{0em}{1pt}{1pt}
\multirow{5}{*}{GPT-4.1}
& Zero-shot CoT & 0.0 & 5.0 & 0.0 & 0.0 & 0.0 & 0.0 & 0.8 \\
& Few-shot CoT {(4-shot)} & 50.0 & 35.0 & 20.0 & 20.0 & 15.0 & 10.0 & 25.0 \\
& ToT & 35.0 & 15.0 &10.0 &5.0 &0.0 &0.0 &10.8\\
& RAP & \textbf{95.0} & 40.0 &10.0 & 0.0 &0.0 &0.0 &24.2\\
& SymPlanner & 75.0 & \textbf{45.0} & \textbf{60.0} & \textbf{50.0} & \textbf{35.0} & \textbf{60.0} & \textbf{54.2}\\
\hline
\end{tabular}}
\end{table}}

\subsection{Main Results}

Table~\ref{tab:main-results} reports symbolic planning success rate across different step counts for GPT-4o-mini, GPT-4o, and GPT-4.1. Success rate is measured as the percentage of generated plans that achieve the goal state under symbolic validation, ensuring that only logically consistent and executable action sequences are counted as correct. 

\paragraph{Scaling with model size.} SymPlanner consistently achieves the highest overall success rate, and the gains become more pronounced with larger base models. For example, SymPlanner reaches $50.0\%$ success rate with GPT-4o and $54.2\%$ with GPT-4.1, compared to only $21.6\%$ with GPT-4o-mini. This scaling trend highlights two important points. First, larger models provide stronger priors for generating semantically meaningful actions, which SymPlanner can further refine through symbolic validation. Second, the improvement gap between small and large models suggests that SymPlanner is capable of exploiting model capacity more effectively than natural language–only baselines, which often plateau or collapse as task difficulty increases. In other words, the framework not only scales with model size but also amplifies the benefits of stronger base LLMs.

\paragraph{Baseline trends.} Among the baselines, few-shot CoT outperforms zero-shot CoT, confirming the benefit of demonstration examples. However, even with four-shot prompting, CoT remains far below SymPlanner, particularly at higher step counts, indicating that prompting improvements alone are insufficient for overcoming error accumulation in long-horizon reasoning. ToT, while more structured than CoT, quickly degrades beyond four steps, reflecting its reliance on breadth-first exploration without an explicit mechanism for grounding or correction. RAP achieves strong results at very short horizons (e.g., $95.0\%$ at two steps with GPT-4.1), but its success rate collapses as depth increases, underscoring its reliance on local step-by-step optimization without global consistency. Collectively, these results demonstrate that current reasoning paradigms lack the mechanisms to effectively constrain or recover from invalid trajectories, making them fragile for complex planning tasks.

\paragraph{Overall insights.} In contrast, SymPlanner sustains competitive performance across long horizons, maintaining $55.0\%$ and $50.0\%$ success rate at eight steps for GPT-4o and GPT-4.1, respectively. The stability of success rate across longer step counts suggests that iterative correction successfully mitigates cascading errors, while contrastive ranking provides reliable selection among competing partial plans. This robustness contrasts sharply with the steep declines observed in baselines, which often fail once reasoning exceeds a handful of steps. Overall, these findings demonstrate that SymPlanner not only improves single-step decision quality but also maintains coherence over extended trajectories, thereby enabling reliable execution in domains that require structured, multi-step planning under strict symbolic constraints. The combination of scalability, robustness, and symbolic grounding positions SymPlanner as a promising framework for bridging natural language reasoning and formal planning systems.

\subsection{Ablation Study}

To assess the contribution of each component in SymPlanner, we conduct an ablation study on GPT-4.1 (Table~\ref{table-ablation}). Success rate is measured as the proportion of generated plans that achieve the goal under symbolic validation. The complete method achieves the highest overall success rate of $54.2\%$, confirming that all three mechanisms are beneficial. 
1) \textbf{Symbolic representation.} Removing the symbolic world model reduces success rate to $46.7\%$. While short-horizon tasks remain solvable, longer trajectories quickly degrade, showing that explicit state tracking is essential for consistency.  
2) \textbf{Iterative correction.} Ablating IC yields the sharpest decline, with success rate falling to $20.8\%$. This confirms IC’s role in preventing single-step errors from compounding, making it the most critical component for long-horizon stability.  
3) \textbf{Contrastive ranking.} Without CR, success rate drops to $27.5\%$, particularly at intermediate horizons. CR enables finer discrimination between superficially similar plans, guiding the search toward coherent trajectories.  
Overall, these results demonstrate that symbolic grounding, correction, and ranking target complementary failure modes and are all necessary for robust, scalable planning.

{\small
\begin{table}[t]
\centering
\caption{Ablation results on GPT-4.1 across different step counts. SymPlanner integrates symbolic representation, iterative correction, and contrastive ranking; each ablation removes one component to evaluate its contribution.}
\label{table-ablation}
\resizebox{0.85\textwidth}{!}{%
\small
\begin{tabular}{ll|ccccccc}
\hline
\specialrule{0em}{1pt}{1pt}
\multirow{2}{*}{\textbf{Model}} & \multirow{2}{*}{\textbf{Method}} & \multicolumn{7}{c}{\textbf{Step Count}}\\
\cmidrule(lr){3-9}
& & 2 & 4 & 6 & 8 & 10 & 12 & Total\\
\hline
\specialrule{0em}{1pt}{1pt}
\multirow{4}{*}{GPT-4.1} 
& Ours & \textbf{75.0} & 45.0 & \textbf{60.0} & \textbf{50.0} & \textbf{35.0} & \textbf{60.0} & \textbf{54.2} \\
& w/o symbolic representation  & 70.0 & \textbf{70.0} & \textbf{60.0} & 40.0 & 30.0 & 10.0 & 46.7 \\
& w/o iterative correction & 50.0 & 35.0  & 10.0 & 20.0 & 0.0 & 10.0 & 20.8 \\
& w/o contrastive ranking & 45.0 & 45.0 & 10.0 & 25.0 & 10.0 & 30.0 & 27.5 \\
\hline
\end{tabular}}
\end{table}}

\section{Related Work}

\paragraph{LLM-based planning.}
Large language models have been applied to planning by generating actions directly in natural language. Some systems incorporate environmental feedback through trial-and-error interactions, building skill memory in specific domains \citep{wang2023describe,wang2023voyager,zhu2023ghost}. These approaches generally target short-horizon tasks and only provide validation after task completion. Other methods, such as ReAct \citep{yao2022react} and RAP \citep{hao2023reasoning}, attempt to simulate state transitions step by step using language. While flexible, such representations are prone to ambiguity and error compounding, leading to unreliable performance in long-horizon settings. SymPlanner differs by explicitly grounding each action in a symbolic world model, which enables deterministic validation and supports iterative correction.

\paragraph{Enhance LLMs with test-time compute.}
Another line of work seeks to improve planning by allocating additional computation during inference. Process reward models (PRMs) \citep{li2024generation,lightman2023let} score intermediate reasoning steps to provide more informative feedback signals for long-horizon reasoning. Although effective in controlled settings, PRMs require costly supervision and often struggle with calibration across diverse tasks \citep{luo2024improve}. SymPlanner instead adopts contrastive ranking, which compares candidate trajectories directly within prompts. This lightweight approach improves discrimination accuracy without the need for additional training or external reward modeling.

\paragraph{Enhance LLMs with symbolic representation.}
Integrating symbolic reasoning with LLMs has emerged as a promising direction for improving robustness and interpretability. Symbolic representations provide structured state descriptions, external knowledge interfaces, and mechanisms for verifying action validity \citep{liu2024evaluating,ni2024comgpt,xiong2024large,yu2024improving,citation-0,he2024give,wang2025taxonomic, valmeekam2023planning}. Within this paradigm, policy strategies can be organized around goal alignment, which evaluates actions by their proximity to the goal, and effect alignment, which validates consistency of predicted outcomes with symbolic rules \citep{kirk2024improving}. SymPlanner follows the effect-alignment perspective by validating actions through a symbolic simulator, while extending prior work with iterative correction to repair invalid actions and contrastive ranking to select coherent long-horizon plans.

\section{Conclusion}

We presented SymPlanner, a deliberate planning framework that augments language models with a symbolic environment to enable structured and verifiable reasoning. By incorporating iterative correction and contrastive ranking, SymPlanner achieves strong performance on PlanBench, demonstrating robustness against error accumulation and improved coherence in long-horizon tasks. Looking forward, an important direction is to integrate reinforcement learning, where symbolic feedback can serve as a reward signal and guide process-aware exploration. Such extensions would allow SymPlanner to autonomously adapt, improve efficiency, and generate correct action sequences in more complex real-world environments.

\bibliographystyle{cogsysapa}
\bibliography{custom}

\newpage

\appendix

\section{Prompts}
In this section, we present all the prompts used in our framework. These prompts include those for action and state generation, action parsing, iterative correction and contrastive ranking. We also include the prompt for plan rating which is used in our ablation study.

\begin{lstlisting}[title=Prompt - Action/State Generation]
I am playing with a set of blocks where I need to arrange the blocks into stacks. Here are the actions I can do

Pick up a block
Unstack a block from on top of another block
Put down a block
Stack a block on top of another block 

I have the following restrictions on my actions:

I can only pick up or unstack one block at a time.
I can only pick up or unstack a block if my hand is empty.
I can only pick up a block if the block is on the table and the block is clear. A block is clear if the block has no other blocks on top of it and if the block is not picked up.
I can only unstack a block from on top of another block if the block I am unstacking was really on top of the other block.
I can only unstack a block from on top of another block if the block I am unstacking is clear.
Once I pick up or unstack a block, I am holding the block.
I can only put down a block that I am holding.
I can only stack a block on top of another block if I am holding the block being stacked.
I can only stack a block on top of another block if the block onto which I am stacking the block is clear.
Once I put down or stack a block, my hand becomes empty.

Use the following FORMAT to solve the task.

### Example Input 1:
"Goal": "The blue block is on the orange block"
"Initial state": "The orange block is clear, the red block is clear, the hand is empty, the orange block is on the blue block, the blue block is on the table, the red block is on the table"

### Example Output 1:
"Action 1": "Unstack the orange block from the blue block"
"State 1": "The blue block is clear, the red block is clear, the hand is holding the orange block, the blue block is on the table, the red block is on the table"
"Action 2": "Put down the orange block on the table"  
"State 2": "The orange block is clear, the blue block is clear, the red block is clear, the hand is empty, the orange block is on the table, the blue block is on the table, the red block is on the table"
"Action 3": "Pick up the blue block"  
"State 3": "The orange block is clear, the red block is clear, the hand is holding the blue block, the orange block is on the table, the red block is on the table"
"Action 4": "Stack the blue block onto the orange block"  
"State 4": "The blue block is on the orange block, the red block is clear, the hand is empty, the orange block is on the table, the red block is on the table, the blue block is clear"
"Goal Achieved": "The blue block is on the orange block"

### Example Input 2:
"Goal": "The blue block is on top of the red block and the yellow block is on top of the orange block"
"Initial state": "The orange block is clear, the yellow block is clear, the hand is empty, the blue block is on top of the red block, the orange block is on top of the blue block, the red block is on the table and the yellow block is on the table"

### Example Output 2:
"Action 1": "Pick up the yellow block"
"State 1": "The orange block is clear, the hand is holding the yellow block, the blue block is on top of the red block, the orange block is on top of the blue block, the red block is on the table"
"Action 2": "Stack the yellow block on top of the orange block"
"State 2": "The yellow block is on top of the orange block, the hand is empty, the blue block is on top of the red block, the orange block is on top of the blue block, the red block is on the table"
"Goal Achieved": "The blue block is on top of the red block and the yellow block is on top of the orange block"

### Example Input 3:
"Goal": "the red block is on top of the blue block and the blue block is on top of the orange block"
"Initial state": "the blue block is clear, the hand is empty, the red block is on top of the yellow block, the blue block is on top of the red block, the yellow block is on top of the orange block and the orange block is on the table"

### Example Output 3:
"Action 1": "Unstack the blue block from the red block"
"State 1": "the blue block is clear, the hand is holding the blue block, the red block is clear, the red block is on top of the yellow block, the yellow block is on top of the orange block, and the orange block is on the table"
"Action 2": "Put down the blue block on the table"
"State 2": "the blue block is clear, the hand is empty, the blue block is on the table, the red block is clear, the red block is on top of the yellow block, the yellow block is on top of the orange block, and the orange block is on the table"
"Action 3": "Unstack the red block from the yellow block"
"State 3": "the red block is clear, the hand is holding the red block, the blue block is clear, the blue block is on the table, the yellow block is clear, the yellow block is on top of the orange block, and the orange block is on the table"
"Action 4": "Put down the red block on the table"
"State 4": "the hand is empty, the red block is clear, the red block is on the table, the blue block is clear, the blue block is on the table, the yellow block is clear, the yellow block is on top of the orange block, and the orange block is on the table"
"Action 5": "Unstack the yellow block from the orange block"
"State 5": "the yellow block is clear, the hand is holding the yellow block, the red block is clear, the red block is on the table, the blue block is clear, the blue block is on the table, the orange block is clear and on the table"
"Action 6": "Put down the yellow block on the table"
"State 6": "the hand is empty, the yellow block is clear, the yellow block is on the table, the red block is clear, the red block is on the table, the blue block is clear, the blue block is on the table, and the orange block is clear and on the table"
"Action 7": "Pick up the blue block"
"State 7": "the hand is holding the blue block, the blue block is clear, the red block is clear, the red block is on the table, the yellow block is clear, the yellow block is on the table, and the orange block is clear and on the table"
"Action 8": "Stack the blue block on top of the orange block"
"State 8": "the hand is empty, the blue block is clear, the blue block is on top of the orange block, the red block is clear, the red block is on the table, the yellow block is clear, the yellow block is on the table, and the orange block is on the table"
"Action 9": "Pick up the red block"
"State 9": "the hand is holding the red block, the red block is clear, the blue block is clear, the blue block is on top of the orange block, the yellow block is clear, the yellow block is on the table, and the orange block is on the table"
"Action 10": "Stack the red block on top of the blue block"
"State 10": "the hand is empty, the red block is on top of the blue block, the blue block is on top of the orange block, the yellow block is clear, the yellow block is on the table, and the orange block is on the table"
"Goal Achieved": "the red block is on top of the blue block and the blue block is on top of the orange block"

### Example Input 4:
"Goal": "the yellow block is on top of the red block, the red block is on top of the blue block and the blue block is on top of the orange block"
"Initial state": "the red block is clear, the yellow block is clear, the hand is empty, the red block is on top of the orange block, the yellow block is on top of the blue block, the blue block is on the table and the orange block is on the table"

### Example Output 4:
"Action 1": "Unstack the yellow block from the blue block"
"State 1": "the red block is clear, the hand is holding the yellow block, the red block is on top of the orange block, the blue block is clear, the blue block is on the table and the orange block is on the table"
"Action 2": "Put down the yellow block on the table"
"State 2": "the yellow block is clear, the hand is empty, the yellow block is on the table, the red block is clear, the red block is on top of the orange block, the blue block is clear, the blue block is on the table and the orange block is on the table"
"Action 3": "Unstack the red block from the orange block"
"State 3": "the hand is holding the red block, the yellow block is clear, the yellow block is on the table, the blue block is clear, the blue block is on the table, the orange block is clear and the orange block is on the table"
"Action 4": "Put down the red block on the table"
"State 4": "the hand is empty, the red block is clear, the red block is on the table, the yellow block is clear, the yellow block is on the table, the blue block is clear, the blue block is on the table and the orange block is clear and on the table"
"Action 5": "Pick up the blue block"
"State 5": "the hand is holding the blue block, the red block is clear, the red block is on the table, the yellow block is clear, the yellow block is on the table and the orange block is clear and on the table"
"Action 6": "Stack the blue block on top of the orange block"
"State 6": "the blue block is clear, the hand is empty, the blue block is on top of the orange block, the red block is clear, the red block is on the table, the yellow block is clear, the yellow block is on the table and the orange block is on the table"
"Action 7": "Pick up the red block"
"State 7": "the hand is holding the red block, the blue block is clear, the blue block is on top of the orange block, the yellow block is clear, the yellow block is on the table and the orange block is on the table"
"Action 8": "Stack the red block on top of the blue block"
"State 8": "the red block is clear, the hand is empty, the red block is on top of the blue block, the blue block is on top of the orange block, the yellow block is clear, the yellow block is on the table and the orange block is on the table"
"Action 9": "Pick up the yellow block"
"State 9": "the hand is holding the yellow block, the red block is clear, the red block is on top of the blue block, the blue block is on top of the orange block and the orange block is on the table"
"Action 10": "Stack the yellow block on top of the red block"
"State 10": "the yellow block is clear, the hand is empty, the yellow block is on top of the red block, the red block is on top of the blue block, the blue block is on top of the orange block and the orange block is on the table"
"Goal Achieved": "the yellow block is on top of the red block, the red block is on top of the blue block and the blue block is on top of the orange block"

\end{lstlisting}

\begin{lstlisting}[title=Prompt - Action Parsing]
I am playing with a set of blocks where I need to arrange the blocks into stacks. Here are the actions I can do

Pick up a block
Unstack a block from on top of another block
Put down a block
Stack a block on top of another block 

Given an action in natural language, I need to parse it into PDDL predicates.

### Example input 1:
pick up the orange block

### Example output 1:
['pickup',  'orange']

### Example input 2:
put down the red block

### Example output 2:
['putdown',  'red']

### Example input 3:
stack the red block on top of the yellow block

### Example output 3:
['stack',  'red', 'yellow']

### Example input 4:
unstack the blue block from on top of the red block

### Example output 4:
['unstack',  'blue', 'red']
\end{lstlisting}

\begin{lstlisting}[title=Prompt - Iterative Correction]
I am playing with a set of blocks where I need to arrange the blocks into stacks. Here are the actions I can do

Pick up a block
Unstack a block from on top of another block
Put down a block
Stack a block on top of another block

I have the following restrictions on my actions:

I can only pick up or unstack one block at a time.
I can only pick up or unstack a block if my hand is empty.
I can only pick up a block if the block is on the table and the block is clear. A block is clear if the block has no other blocks on top of it and if the block is not picked up.
I can only unstack a block from on top of another block if the block I am unstacking was really on top of the other block.
I can only unstack a block from on top of another block if the block I am unstacking is clear.
Once I pick up or unstack a block, I am holding the block.
I can only put down a block that I am holding.
I can only stack a block on top of another block if I am holding the block being stacked.
I can only stack a block on top of another block if the block onto which I am stacking the block is clear.
Once I put down or stack a block, my hand becomes empty.

Now I need to correct previous invalid actions. Use the following FORMAT.


### Example Input 1:
"Goal": "The blue block is on the orange block"
"Initial state": "The orange block is clear, the red block is clear, the hand is empty, the orange block is on the blue block, the blue block is on the table, the red block is on the table"
"Action 1": "Unstack the orange block from the blue block"
"State 1": "The blue block is clear, the red block is clear, the hand is holding the orange block, the blue block is on the table, the red block is on the table"
"Previous invalid actions": {"Action 2": "Pick up the red block"}

### Example Output 1:
"Action 2": "Put down the orange block"

### Example Input 2:
"Goal": "The blue block is on top of the red block and the yellow block is on top of the orange block"
"Initial state": "The orange block is clear, the yellow block is clear, the hand is empty, the blue block is on top of the red block, the orange block is on top of the blue block, the red block is on the table and the yellow block is on the table"
"Previous invalid actions": {"Action 1": "Unstack the blue block from the red block", "Action 1": "Pick up the blue block", "Action 1": "Pick up the red block"}

### Example Output 2:
"Action 1": "Pick up the yellow block"

### Example Input 3:
"Goal": "The blue block is on top of the orange block"
"Initial state": "The red block is clear, the yellow block is clear, the hand is empty, the red block is on top of the blue block, the yellow block is on top of the orange block, the blue block is on the table and the orange block is on the table"
"Previous invalid actions": {"Action 1": "Unstack the blue block from the red block", "Action 1": "Pick up the blue block", "Action 1": "Put down the yellow block"}

### Example Output 3:
"Action 1": "Unstack the yellow block from the orange block"

### Example Input 4:
"Goal": "The red block is on top of the orange block"
"Initial state": "The orange block is clear, the yellow block is clear, the hand is empty, the blue block is on top of the red block, the orange block is on top of the blue block, the red block is on the table and the yellow block is on the table"
"Action 1": "Unstack the orange block from the blue block"
"State 1": "the yellow block is clear, the blue block is clear, the hand is holding the orange block, the blue block is on top of the red block, the red block is on the table and the yellow block is on the table"
"Action 2": "Put down the orange block"
"State 2": "the yellow block is clear, the orange block is clear, the blue block is clear, the hand is empty, the blue block is on top of the red block, the red block is on the table, the yellow block is on the table and the orange block is on the table"
"Previous invalid actions": {"Action 3": "Put down the orange block", "Action 3": "Pick up the red block"}

### Example Output 4:
"Action 3": "Pick up the yellow block"
\end{lstlisting}

\begin{lstlisting}[title=Prompt - Contrastive Ranking]
You will be given a goal, an initial block configuration, one or more candidate actions, and their resulting future states. Your task is to evaluate the options by considering both their immediate effects and their utility toward achieving the stated goal.
Follow this exact format in your response:
1. "Comparison": Provide a detailed analysis of each option. Evaluate how well each action adheres to the rules, affects the current state, and contributes to achieving the goal. If applicable, consider the long-term impact of the action in reaching the final configuration. 
2. "Conclusion": Choose only one option that is most effective in progressing toward the goal. Your answer must be in the form: "Option 1" or "Option 2" or "Option 3" - no additional explanation should be included in this section.


### Example Input 1:
"Goal": "The blue block is on the orange block",
"Initial state": "The orange block is clear, the red block is clear, the hand is empty, the orange block is on the blue block, the blue block is on the table, the red block is on the table",
"Search steps": {
    "Option 1": {"Action 1": "Unstack the orange block from the blue block"},
    "Option 2": {"Action 1": "Pickup the red block"}
},
"Futures": {
    "Future 1": {"State 1": "The blue block is clear, the red block is clear, the hand is holding the orange block, the blue block is on the table, the red block is on the table"},
    "Future 2": {"State 1": "The orange block is clear, the hand is holding the red block, the orange block is on the blue block, the blue block is on the table"}
}

### Example Output 1:
"Comparison": "Option 1 adheres to the rules. This is a productive move toward achieving the goal "The blue block is on the orange block" because now the orange block can be placed on the table, and then the blue block can be stacked on it. Option 2 is also valid. However, it does not progress toward achieving the goal, as it neither changes the relationship between the blue and orange blocks nor facilitates doing so in subsequent steps.",
"Conclusion": "Option 1"


### Example Input 2:
"Goal": "The blue block is on top of the red block and the yellow block is on top of the orange block",
"Initial state": "The orange block is clear, the yellow block is clear, the hand is empty, the blue block is on top of the red block, the orange block is on top of the blue block, the red block is on the table and the yellow block is on the table",
"Action 1": "Pick up the yellow block",
"State 1": "The orange block is clear, the hand is holding the yellow block, the blue block is on top of the red block, the orange block is on top of the blue block, the red block is on the table",
"Search steps": {
    "Option 1": {"Action 2": "Put down the yellow block on the table"},
    "Option 2": {"Action 2": "Stack the yellow block on top of the orange block"}
},
"Futures": {
    "Future 1": {"State 2": "The orange block is clear, the hand is empty, the blue block is on top of the red block, the orange block is on top of the blue block, the red block is on the table, the yellow block is on the table"},
    "Future 2": {"State 2": "The yellow block is on top of the orange block, the hand is empty, the blue block is on top of the red block, the orange block is on top of the blue block, the red block is on the table"}
}

### Example Output 2:
"Comparison": "Option 1 places the yellow block back on the table, which does not contribute to the goal of stacking it on the orange block. It keeps the yellow block isolated and results in a non-progressive state. Option 2 stacks the yellow block directly on top of the orange block, achieving the goal, as one half of the desired configuration (yellow on orange) is satisfied. Additionally, Option 2 does not interfere with the already correct configuration of blue on red. Therefore, Option 2 is the only one that moves toward achieving the full goal state.",
"Conclusion": "Option 2"


### Example Input 3:
"Goal": "the red block is on top of the blue block and the blue block is on top of the orange block",
"Initial state": "the blue block is clear, the hand is empty, the red block is on top of the yellow block, the blue block is on top of the red block, the yellow block is on top of the orange block and the orange block is on the table",
"Action 1": "unstack the blue block from on top of the red block",
"State 1": "the red block is clear, the hand is holding the blue block, the red block is on top of the yellow block, the yellow block is on top of the orange block, and the orange block is on the table",
"Action 2": "put down the blue block",
"State 2": "the red block is clear, the blue block is clear, the hand is empty, the red block is on top of the yellow block, the yellow block is on top of the orange block, the blue block is on the table and the orange block is on the table",
 "Search steps": {
    "Option 1": {"Action 3": "unstack the red block from on top of the yellow block"},
    "Option 2": {"Action 3": "pick up the blue block"}
},
"Futures": {
    "Future 1": {"State 3": the blue block is clear, the yellow block is clear, the hand is holding the red block, the yellow block is on top of the orange block, the blue block is on the table and the orange block is on the table"},
    "Future 2": {"State 3": "the red block is clear, the hand is holding the blue block, the red block is on top of the yellow block, the yellow block is on top of the orange block and the orange block is on the table"}
}

### Example Output 3:
"Comparison": "Option 1 unstacking the red block from on top of the yellow block is a productive step toward the goal. It frees the yellow block and allows the red block to be repositioned, which is necessary because ultimately the red block needs to be on top of the blue block, and the blue block is currently on the table. This action sets up the possibility of stacking the blocks in the correct order. Option 2, picking up the blue block, simply returns to a previous state where the blue block is being held, but does not address the need to clear the yellow block or move the red block. Therefore, Option 1 is more effective in progressing toward the goal configuration.",
"Conclusion": "Option 1"


### Example Input 4:
"Goal": "the yellow block is on top of the red block, the red block is on top of the blue block and the blue block is on top of the orange block",
"Initial state": "the red block is clear, the yellow block is clear, the hand is empty, the red block is on top of the orange block, the yellow block is on top of the blue block, the blue block is on the table and the orange block is on the table",
"Action 1": "unstack the yellow block from on top of the blue block",
"State 1": "the red block is clear, the blue block is clear, the hand is holding the yellow block, the red block is on top of the orange block and the blue block is on the table",
"Action 2": "put down the yellow block",
"State 2": "the red block is clear, the blue block is clear, the yellow block is clear, the hand is empty, the red block is on top of the orange block, the blue block is on the table, the orange block is on the table and the yellow block is on the table",
"Action 3": "unstack the red block from on top of the orange block",
"State 3": "the orange block is clear, the blue block is clear, the yellow block is clear, the hand is holding the red block, the blue block is on the table, the orange block is on the table and the yellow block is on the table",
"Action 4": "put down the red block",
"State 4": "the orange block is clear, the red block is clear, the blue block is clear, the yellow block is clear, the hand is empty, the orange block is on the table, the red block is on the table, the blue block is on the table and the yellow block is on the table",
 "Search steps": {
    "Option 1": {"Action 5": "stack the red block on top of the yellow block"},
    "Option 2": {"Action 5": "stack the orange block on top of the blue block"},
    "Option 3": {"Action 5": "pick up the blue block"},
},
"Futures": {
    "Future 1": {"State 5": "the orange block is clear, the blue block is clear, the yellow block is clear, the hand is empty, the orange block is on the table, the blue block is on the table and the yellow block is on the table"},
    "Future 2": {"State 5": "the red block is clear, the blue block is clear, the yellow block is clear, the hand is empty, the orange block is on the table, the red block is on the table, the blue block is on the table and the yellow block is on the table"},
    "Future 3": {"State 5": "the orange block is clear, the red block is clear, the yellow block is clear, the hand is holding the blue block, the orange block is on the table, the red block is on the table and the yellow block is on the table"},
}

### Example Output 4:
"Comparison": "Option 1 attempts to stack the red block on top of the yellow block, but according to State 4, the hand is empty and the red block is on the table, so this action is invalid as the hand must be holding the red block to stack it. Option 2 attempts to stack the orange block on top of the blue block, but similarly, the hand is empty and the orange block is on the table, so this action is also invalid. Option 3 picks up the blue block, which is a valid action since the hand is empty and the blue block is clear and on the table. This action is productive because, for the goal state, the blue block needs to be stacked on top of the orange block, and picking it up is a necessary first step toward that configuration. Therefore, Option 3 is the only valid and productive move toward achieving the goal.",
"Conclusion": "Option 3"

\end{lstlisting}

\begin{lstlisting}[title=Prompt - Plan Rating]
You will be provided with a goal, an initial block configuration, and a partial plan. Your task is to evaluate the quality of the partial plan based on:
- Whether the actions are valid and maintain consistency with the preconditions and postconditions.
- Whether the actions help achieve the goal, either directly or by setting up future beneficial moves.

You should DIRECTLY give a score from 1 to 10:
- 1: Invalid or counterproductive actions.
- 5: Neutral actions - valid but not helpful toward the goal.
- 10: Highly effective actions - valid and clearly beneficial for reaching the goal. 

## Example 1:
### Input:
"Goal": "The blue block is on the orange block"
"Initial state": "The orange block is clear, the red block is clear, the hand is empty, the orange block is on the blue block, the blue block is on the table, the red block is on the table"

### Output:
"Action 1": "Unstack the orange block from the blue block"
"State 1": "The blue block is clear, the red block is clear, the hand is holding the orange block, the blue block is on the table, the red block is on the table"

### Rating
8

## Example 2:
### Input:
"Goal": "The blue block is on top of the red block and the yellow block is on top of the orange block"
"Initial state": "The orange block is clear, the yellow block is clear, the hand is empty, the blue block is on top of the red block, the orange block is on top of the blue block, the red block is on the table and the yellow block is on the table"

### Output:
"Action 1": "Pick up the yellow block"
"State 1": "The orange block is clear, the hand is holding the yellow block, the blue block is on top of the red block, the orange block is on top of the blue block, the red block is on the table"
"Action 2": "Put down the yellow block on the table"
"State 2": "The orange block is clear, the hand is empty, the blue block is on top of the red block, the orange block is on top of the blue block, the red block is on the table, the yellow block is on the table"

### Rating:
3
\end{lstlisting}

\end{document}